\documentclass{article}

\usepackage[preprint]{neurips_2025}

\usepackage[utf8]{inputenc} 
\usepackage[T1]{fontenc}    
\usepackage{hyperref}       
\usepackage{url}            
\usepackage{booktabs}       
\usepackage{amsfonts}       
\usepackage{nicefrac}       
\usepackage{microtype}      
\usepackage{xcolor}         
\usepackage{amsmath}
\usepackage{amssymb}
\usepackage{amsthm}
\usepackage{marvosym}
\usepackage{algorithm}
\usepackage{algpseudocode}
\usepackage{graphicx}
\usepackage{subcaption}
\theoremstyle{definition}
\newtheorem{definition}{Definition}

\title{Reinforced Context Order Recovery for Adaptive Reasoning and Planning}

\author{
  Long Ma \\
  Peking University \\
  Beijing, China \\
  \texttt{malong@pku.edu.cn} \\
  \And
  Fangwei Zhong\textsuperscript{\Letter} \\
  Beijing Normal University\\
  Beijing, China \\
  \texttt{fangweizhong@bnu.edu.cn} \\
  \And
  Yizhou Wang\\
  Peking University \\
  Beijing, China \\
  \texttt{yizhou.wang@pku.edu.cn} \\
}

\begin{document}

\maketitle

\begin{abstract}
Modern causal language models, followed by rapid developments in discrete diffusion models, can now produce a wide variety of interesting and useful content.
However, these families of models are predominantly trained to output tokens with a fixed (left-to-right) or random order, which may deviate from the logical order in which tokens are generated originally.
In this paper, we observe that current causal and diffusion models encounter difficulties in problems that require adaptive token generation orders to solve tractably, which we characterize with the $\mathcal{V}$-information framework.
Motivated by this, we propose Reinforced Context Order Recovery (ReCOR), a reinforcement-learning-based framework to extract adaptive, data-dependent token generation orders from text data without annotations.
Self-supervised by token prediction statistics, ReCOR estimates the hardness of predicting every unfilled token and adaptively selects the next token during both training and inference.
Experiments on challenging reasoning and planning datasets demonstrate the superior performance of ReCOR compared with baselines, sometimes outperforming oracle models supervised with the ground-truth order.
\end{abstract}

\section{Introduction}\label{sec:intro}

Text generation models have seen remarkable advancements in the past few years, with causal language models (CLMs) trained by next-token prediction taking the lead~\cite{achiam2023gpt,touvron2023llama,team2023gemini}, followed by more recent discrete diffusion models~\cite{hoogeboom2021argmax,austin2021structured,nie2025llada}.
These classes of models have demonstrated impressive capabilities across a wide range of tasks, from writing code to executing tasks as agents~\cite{wei2022cotreasoning,wang2024llmagent,guo2024deepseek}.

\begin{figure}[t]
  \centering
  \includegraphics[width=\textwidth]{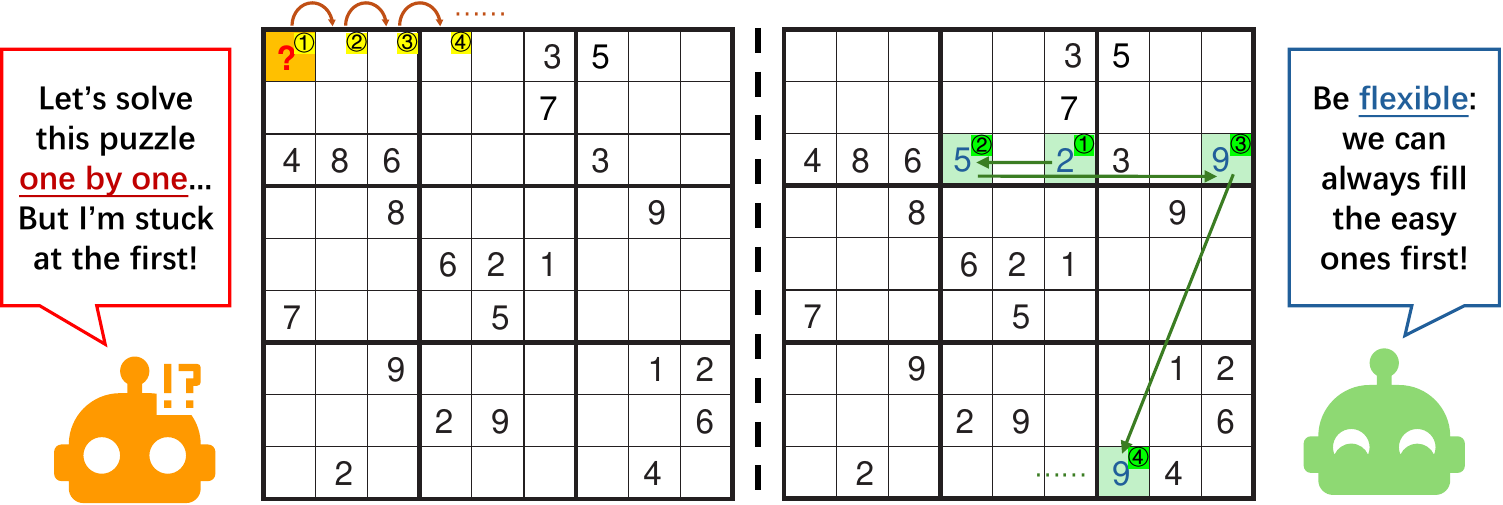}
  
  \caption{Illustration of ReCOR (right) compared with standard causal language modeling (left). While causal language modeling always tries to produce tokens left-to-right, ReCOR estimates the hardness of each token and adaptively prioritizes the easy ones without external supervision.}
  \label{fig:illustration}
  \vspace{-0.2cm}
\end{figure}

Despite the developments, current models still encounter significant challenges when faced with complex reasoning and planning problems that require a long-horizon and flexible decision-making process.
A crucial part of this challenge lies in the token generation order~\cite{shah2024causal}.
As illustrated in Fig.~\ref{fig:illustration}, in adaptive reasoning tasks like Sudoku, some cells could be hard to predict instantly, depending on other cells to be filled first and provide additional constraints to eliminate candidates.
However, CLMs always follow a rigid left-to-right generation paradigm, encountering many intractable tokens along the way.
In contrast, humans rarely solve complex reasoning problems in a strictly linear fashion; we tackle the easiest parts first and use those insights to address progressively more challenging ones.
Taking note of this issue, existing works either explicitly train the model to predict the easiest next token~\cite{shah2024causal} or leverage the properties of masked diffusion models (MDMs)~\cite{ye2025beyond,kim2025train} to adaptively decode easy tokens during inference.
However, these methods require additional annotations or introduce large distribution shifts between training and inference, hurting performance.

Facing these challenges, in this paper, we introduce Reinforced Context Order Recovery (ReCOR), a self-supervised framework that learns to adaptively determine the optimal token generation order without explicit order annotations.
Our approach is motivated by the insight that the hardness of predicting different tokens varies dramatically conditioned on the current context, which we characterize using the framework of predictive $\mathcal{V}$-information.
To operationalize the objective derived under this framework, ReCOR casts order prediction as a decision-making problem and trains a policy that adaptively selects the next token (Fig.~\ref{fig:illustration}).
ReCOR jointly optimizes the token prediction model and the order prediction policy, generating rewards with the former as self-supervision for the latter.
Furthermore, unlike previous approaches that apply adaptive strategies for inference only, ReCOR follows the same distribution of order during both training and inference.
This ensures that the model not only adapts its generation behavior during inference but also benefits from learning informative, tractable token prediction tasks during training.
In the experiments, we demonstrate the effectiveness of ReCOR on a variety of reasoning and planning tasks, including arithmetic problems and logic puzzles, where ReCOR consistently outperforms the state-of-the-art methods.

The contributions of our paper are fourfold:
1) We observe and characterize the token ordering problem using the framework of predictive $\mathcal{V}$-information and propose a corresponding objective;
2) We propose an RL-based formulation and algorithm for optimizing the objective and obtaining an adaptive order predictor;
3) We empirically validate ReCOR on multiple reasoning and planning benchmarks, demonstrating state-of-the-art performance that is competitive or even better than oracle models with access to ground-truth orders;
4) We offer an analysis on specific failure modes of inference-only adaptive methods, showing the necessity of our training-time adaptive designs.

\vspace{-0.2cm}
\section{Related Work}\label{sec:related_work}
\vspace{-0.2cm}

\textbf{(Any-order) Autoregressive Models and (Discrete) Diffusion Models.}
Autoregressive and diffusion models are currently the two dominant families of generative models in various domains~\cite{yang2023diffusionsurvey,cao2024gendiffusionsurvey,xiong2024autoregressivevisionsurvey}.
For textual modeling, in recent years, there has been a seismic rise in popularity for large language models~\cite{achiam2023gpt,touvron2023llama,team2023gemini} of which the vast majority are trained using the next-token prediction objective, or \textit{causal language modeling}, e.g. GPT~\cite{radford2019gpt2,brown2020language}.
Alternatively, discrete variants~\cite{austin2021structured,hoogeboom2021argmax} of diffusion models~\cite{DDPM,song2019generative,song2021scorebased} like masked diffusion models (MDMs) apply a denoising objective to randomly corrupted data to learn the underlying structures.
Furthermore, any-order autoregressive models (AO-ARMs)~\cite{yang2019xlnet, shih2022therightway} generalize the causal modeling paradigm by allowing arbitrary orders or masks during training and inference.
Compared with these approaches, ReCOR uses a learned adaptive order during both training and inference, avoiding issues with fixed or random orders.
We primarily compare to adaptive inference variants of MDMs~\cite{ye2025beyond,kim2025train} since AO-ARMs are trained similarly.

\textbf{Token Ordering and Reflections on Causal Language Modeling.}
Within the active research field of language models, there have long been criticisms and attempts at improvements regarding various aspects of the causal modeling paradigm, including autoregressive inference~\cite{dziri2023faith}, teacher-forcing training~\cite{bachmann2024pitfalls}, and the left-to-right order~\cite{du2023autoregressive}.
In particular, current LLMs have been found to struggle with many complex reasoning and planning problems, especially adaptive ones~\cite{valmeekam2023planning,shah2024causal,kambhampati2024llmcantplan,li2024cotserial}.
We focus on token ordering, motivated by the fact that these reasoning problems often require intricate, data-dependent orders to be tractably solved.
Recently, adaptive inference methods based on MDMs~\cite{ye2025beyond,kim2025train} were proposed to address this problem, combining random masking at training time with selective token inference.
We note that adaptive orders are needed during both training and inference, and design ReCOR to automatically recover the correct order with self-supervision.

\section{Preliminary}\label{sec:prelim}

\subsection{Problem Setup}\label{sec:problem_setup}

We focus our attention on the classic sequence-to-sequence setting, where the goal is to learn a conditional distribution $p(\mathbf{y} \mid \mathbf{x})$ with $\mathbf{x} \in \mathcal{X} \subseteq \mathcal{T}^*$ as the prompt (space), $\mathbf{y} \in \mathcal{Y} \subseteq \mathcal{T}^*$ as the response (space), and $\mathcal{T}$ as the token vocabulary.
For simplicity, we assume prompts of length $N$ and responses of length $M$: $\mathbf{x}=(x_1, x_2, \ldots, x_N) \in \mathcal{X} =\mathcal{T}^N, \mathbf{y}=(y_1, y_2, \ldots, y_M) \in \mathcal{Y} =\mathcal{T}^M$ which can be achieved without loss of generality through padding.
A training dataset $D=\{(\mathbf{x},\mathbf{y})\}$ is available with i.i.d. samples $(\mathbf{x}, \mathbf{y}) \sim p(\mathbf{x})p(\mathbf{y} \mid \mathbf{x})$.

\subsection{Existing Approaches}\label{sec:existing_approaches}

In recent years, \textbf{causal language models}~(CLMs) have witnessed a giant wave of interest, trained using the prevalent \textbf{next-token prediction}~(NTP) objective:
\begin{equation}\label{eq:ntp}
    \min_\theta \mathbb{E}_{(\mathbf{x}, \mathbf{y}) \sim D} \left[-\sum_{i=1}^M \log p_\theta(y_i \mid \mathbf{x}, \mathbf{y}_{<i}) \right]
\end{equation}
with $p_\theta(\cdot \mid \mathbf{x}, \mathbf{y}_{<i})$ parameterized as an autoregressive sequence model (e.g. GPT~\cite{radford2019gpt2,brown2020language}).

As a generalization of CLMs, \textbf{any-order autoregressive models}~(AO-ARMs)~\cite{yang2019xlnet,shih2022therightway} perform "next-token" predictions under an arbitrary generation order $\boldsymbol{\rho} \in S_M$ instead of the causal order:
\begin{equation}\label{eq:aoarm}
    \min_\theta \mathbb{E}_{(\mathbf{x}, \mathbf{y}) \sim D, \boldsymbol{\rho} \sim \mathcal{U}_{\boldsymbol{\rho}}} \left[ -\sum_{i=1}^M \log p_\theta(y_{\rho_i} \mid \mathbf{x}, \mathbf{y}_{\boldsymbol\rho_{<i}}, \rho_i) \right]
\end{equation}
where $\mathcal{U}_{\boldsymbol\rho}$ is the distribution of training orders, usually taken as the uniform distribution over all $M$-permutations $S_M$~\cite{yang2019xlnet} or further canonicalized using simple rules~\cite{shih2022therightway}.
Note that $\boldsymbol\rho$ is the \textbf{generation order} and distinct from the raw \textbf{content order}, which is still retained; i.e. for $\mathbf{y}=\verb|abc|$ and $\rho_1=3$, the partially generated response at the first step should be \verb|**c| instead of \verb|c**|.

Alternatively, \textbf{masked diffusion models} (MDMs)~\cite{austin2021structured,hoogeboom2021argmax,lou2024discrete,kim2025train} replace some response tokens with a special \verb|[MASK]| token in a forward process $q_{t \mid 0}(\mathbf{y}^t \mid \mathbf{y})$ that masks each token independently at random, then learn to predict the masked tokens with a denoising objective (constants omitted):
\begin{equation}\label{eq:mdm}
    \min_\theta \mathbb{E}_{(\mathbf{x}, \mathbf{y}) \sim D, t \sim \mathcal{U}[0, 1], \mathbf{y}^t \sim q_{t \mid 0}(\cdot \mid \mathbf{y})} \left[ -\sum_{i=1}^M \mathbb{I}[y_i^t=\verb|[MASK]|] \log p_\theta(y_i \mid \mathbf{x}, \mathbf{y}^t, t, i) \right]
\end{equation}
which is also often implemented with a transformer.

\section{Problem Analysis and Method}\label{sec:method}

This section is structured as follows:
We first motivate our focus on generation order by characterizing \textit{token hardness} (Sec.~\ref{sec:token_hardness}) and propose to cast order recovery as a decision-making problem (Sec.~\ref{sec:decision_making}).
The core training algorithm for ReCOR is presented in Sec.~\ref{sec:alg} with architectural details in Sec.~\ref{sec:arch}.

\subsection{Generation Order and Token Hardness}\label{sec:token_hardness}
We begin our discussion about the order problem by noting that prior works mostly handle generation orders passively during training, delegating the issue to a fixed data-independent strategy (Sec.~\ref{sec:existing_approaches}, always left-to-right for CLM and uniformly random for AO-ARM and MDM).
This is justified from a probabilistic standpoint, where any joint distribution over a series of random variables can be decomposed arbitrarily into the product of a series of conditional distributions via the chain rule: $P(y_1, y_2, \ldots, y_M)=\prod_{i=1}^M P(y_{\rho_i} \mid y_{\rho_1}, \ldots, y_{\rho_{i-1}}), \forall \boldsymbol{\rho} \in S_M$.
Furthermore, for problems with a unique solution, the prompt can completely determine each response token, and we have $H(Y_{\rho_i} \mid X, \{Y_{\rho_j}\}_{j <i}) \leq H(Y_{\rho_i}\mid X)=0, I(Y_{\rho_i}; \{Y_{\rho_j}\}_{j < i} \mid X)=0, \forall \boldsymbol{\rho},i$, so predicted response tokens do not bring additional information, also implying that order is irrelevant.

However, as illustrated in Fig.~\ref{fig:illustration} and observed by prior works~\cite{shah2024causal,kim2025train,ye2025beyond}, for hard reasoning and planning problems, plain causal modelling frequently encounters computationally intractable intermediate steps and fails dramatically; a tailored generation order is required for each instance to make the problem tractable in practice.
Consequently, we choose to explicitly model the generation order $\boldsymbol{\rho}$ as an unobserved variable to be recovered from the plain-text-only training data.

Under such a formulation, the problem then turns to defining and finding a \textit{good} $\boldsymbol\rho$.
Intuitively, we would like to start with the easy parts of the response and work our way to the harder parts step-by-step, utilizing the partially generated solution as a form of chain-of-thought to guide later generations.
We formalize this intuition using the framework of predictive $\mathcal{V}$-information~\cite{Xu2020vinfo}:
\begin{definition}[Predictive $\mathcal{V}$-information~\cite{Xu2020vinfo}]\label{def:pvi}
    Let $\mathcal{V} \subseteq \{f: \mathcal{A} \cup \{\varnothing\} \rightarrow \Delta_\mathcal{B}\}$ be a \textbf{predictive function class} that predicts the distribution of a random variable taking values over $\mathcal{B}$, optionally given the value of another random variable over $\mathcal{A}$ under some regularity conditions.
    
    Define the \textbf{predictive conditional $\mathcal{V}$-entropy} of random variables $A, B$ as
    
    \begin{align}
        H_\mathcal{V}(B \mid A) &= \inf_{f \in \mathcal{V}} \mathbb{E}_{a,b \sim A, B}[-\log f[a](b)] \\
        H_\mathcal{V}(B \mid \varnothing) &= \inf_{f \in \mathcal{V}} \mathbb{E}_{b \sim B}[-\log f[\varnothing](b)]
    \end{align}
    
    Then the \textbf{predictive $\mathcal{V}$-information} from random variable $A$ to $B$ is defined as
    \begin{equation}
        I_\mathcal{V}(A \rightarrow B)=H_\mathcal{V}(B \mid \varnothing) - H_\mathcal{V}(B \mid A).
    \end{equation}
\end{definition}

Def.~\ref{def:pvi} captures the \textit{hardness} of a token given the current context under computational constraints, which evades standard probabilistic and information-theoretic arguments.
Given prompt $X$ and a set of already predicted response tokens $\{Y_{\rho_j}\}_{j < i}$, an easy token $Y_{\rho_i}$ would have a large $I_\mathcal{V}(X, \{Y_{\rho_j}\}_{j<i} \rightarrow Y_{\rho_i})$ whereas a hard token would have a small $\mathcal{V}$-information.
Building on this characterization, we set our overall objective to maximize the following cumulative predictive $\mathcal{V}$-information using a learned autoregressive $\boldsymbol{\rho}$-generator parameterized by $\theta$:
\begin{equation}
    \max_\theta \sum_{i=1}^M \mathbb{E}_{\rho_i \sim p_\theta(\cdot \mid X, \{Y_{\rho_j}\}_{j<i})} I_\mathcal{V}(X,\{Y_{\rho_j}\}_{j<i} \rightarrow Y_{\rho_i})
\end{equation}
Note that Def.~\ref{def:pvi} of $\mathcal{V}$-information contains an optimization problem.
Naively training a new predictor to solve the optimization problem within $I_\mathcal{V}(X,\{Y_{\rho_j}\}_{j<i} \rightarrow Y_{\rho_i})$ for every $\boldsymbol{\rho}_{\leq i}$ would require an exponential number of training instances.
For tractability, we set the predictive function class $\mathcal{V}$ to the family of autoregressive token generators conditioned on arbitrary contexts and parameterized by $\psi$, and share it across inner subproblems for all $\boldsymbol{\rho}$, yielding the following objective up to constants:
\begin{equation}\label{eq:main_obj}
    \max_{\theta, \psi} \mathbb{E}_{(\mathbf{x}, \mathbf{y}) \sim D} \left[\sum_{i=1}^M \mathbb{E}_{\rho_i \sim p_\theta(\cdot \mid \mathbf{x}, \mathbf{y}_{\boldsymbol{\rho}_{<i}})} \log p_\psi(y_{\rho_i} \mid \mathbf{x}, \mathbf{y}_{\boldsymbol{\rho}_{<i}}, \rho_i) \right]
\end{equation}

\subsection{Order Recovery as a Decision-making Problem}\label{sec:decision_making}

\begin{figure}[t]
  \centering
  \includegraphics[width=\textwidth]{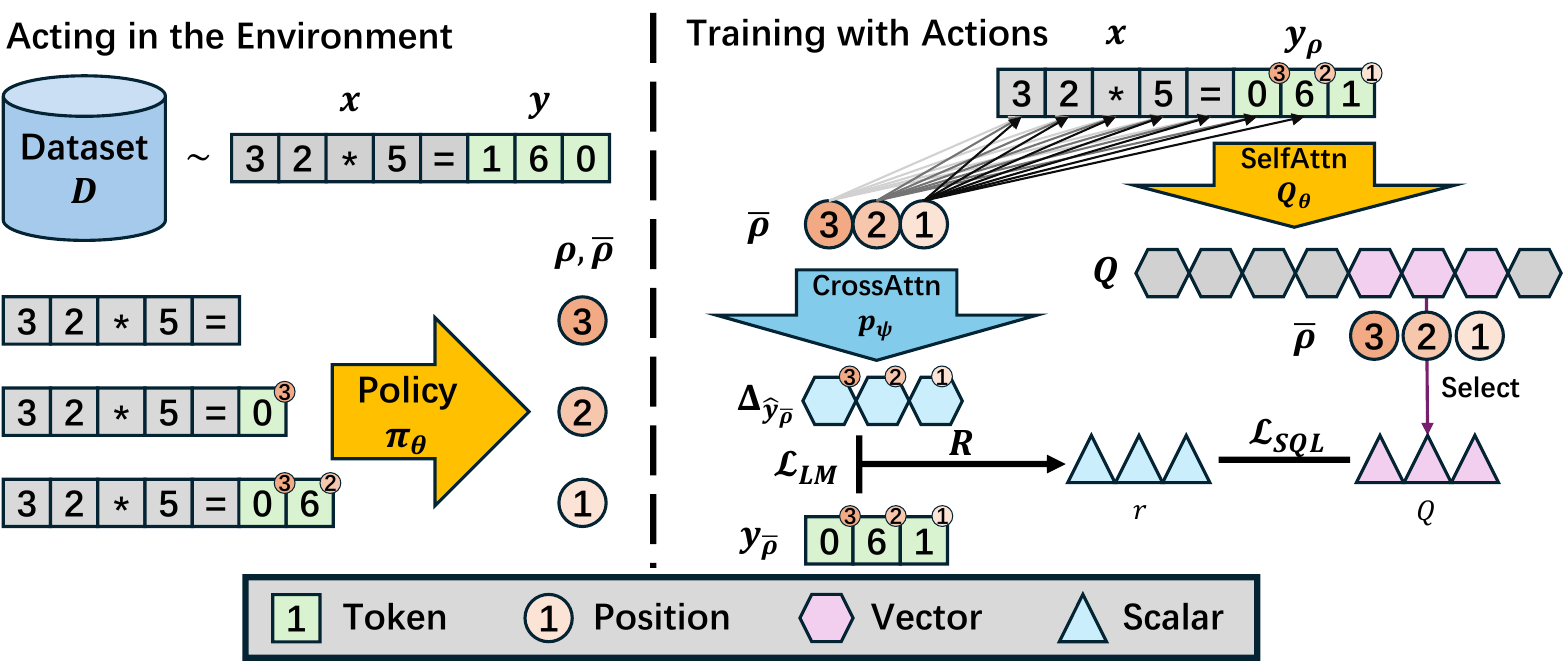}
  
  \caption{Illustration of the training procedure of ReCOR. ReCOR first rolls out the order prediction policy $\pi_\theta$ autoregressively on sampled data $(\mathbf{x}, \mathbf{y})$ to obtain the actions $\boldsymbol{\rho},\bar{\boldsymbol{\rho}}$ (left), then performs parallelized training on the sampled actions (right). The token predictor $p_\psi$ is trained to answer queries generated by $\pi_\theta$ using $\mathcal{L}_\text{LM}$ while $Q_\theta$ is supervised by reward signals from $p_\psi$ using $\mathcal{L}_\text{SQL}$.}
  \label{fig:method}
\end{figure}

We now proceed to solve Eq.~\ref{eq:main_obj} for $\theta$ and the associated $\psi$.
Although \cite{Xu2020vinfo} proposed a method for structure learning based on the Chu-Liu algorithm~\cite{chu1965shortest}, it assumes that the edge weight depends only on the two vertices connected by the edge, while under our setting, the likelihood depends on all random variables in the context with an exponential number of combinations.
Furthermore, structure learning requires a uniform structure over the whole dataset, while the correct generation order can be data-dependent and vary between instances (e.g. Sudoku~\cite{shah2024causal} in Fig.~\ref{fig:illustration}).

Facing these challenges, we formulate the recovery of $\boldsymbol{\rho}$ as a decision-making problem and employ reinforcement learning (RL) techniques~\cite{sutton2018reinforcement,haarnoja2017softql,schulman2017proximal} to train the $\rho$-generator.
We construct the following Markov decision process (MDP) $(\mathcal{S}, \mathcal{A}, P, R, p_1, \gamma)$ where $\mathcal{S}:=\mathcal{X} \times \cup_{I\subseteq [M]} \mathcal{T}^I$ is the state space with the prompt and a partially generated response;
$\mathcal{A}:=[M]$ is the action space corresponding to the position of the next token to generate;
$P$ is the transition that adds the newly generated token to the state;
$R$ is the reward function as in Eq.~\ref{eq:reward_ppl} and \ref{eq:reward_sparse};
$p_1:=p(\mathbf{x}) \times \delta(\varnothing)$ is the initial state distribution consisting of a sampled prompt and an empty response;
and $\gamma \in [0, 1]$ is the discount factor.
Concretely, at time step $1 \leq t \leq M$, $s_t = (\mathbf{x}, \mathbf{y}_{\boldsymbol{\rho}_{<t}}),a_t=\rho_t$, which transitions to $(\mathbf{x}, \mathbf{y}_{\boldsymbol{\rho}_{\leq t}})$ with $y_{\rho_t}$ taken from the training data or sampled from $p_\psi(\cdot \mid \mathbf{x}, \mathbf{y}_{\boldsymbol{\rho}_{<t}}, \rho_t)$ during inference.
Under this formulation, the objective is to train a policy $\pi_\theta: \mathcal{S} \rightarrow \Delta_\mathcal{A}$ satisfying
\begin{equation}\label{eq:rl}
    \max_\theta \mathbb{E}_{s_1 \sim p_1, a_t \sim \pi_\theta(\cdot \mid s_t), r_t \sim R(s_t, a_t), s_{t+1} \sim P(s_t, a_t)} \left[ \sum_t \gamma^t r_t \right]
\end{equation}
which recovers Eq.~\ref{eq:main_obj} with $\pi_\theta(\cdot \mid s_t)=p_\theta(\cdot \mid \mathbf{x}, \mathbf{y}_{\boldsymbol{\rho}_{<t}}),R(s_t, a_t)=\log p_\psi(y_{\rho_t} \mid \mathbf{x}, \mathbf{y}_{\boldsymbol{\rho}_{<t}}, \rho_t)$ and $\gamma=1$.
This formulation also allows more room for design choices, e.g. RL algorithms, discount factor, and alternative reward functions; see Sec.~\ref{sec:alg} below for details.

Equipped with $\pi_\theta$ for order prediction and $p_\psi$ for token prediction, we can now perform inference by sampling $\rho_t \sim \pi_\theta(\cdot \mid \mathbf{x}, \hat{\mathbf{y}}_{\boldsymbol{\rho}_{<t}})$ and $\hat{y}_{\rho_t} \sim p_\psi(\cdot \mid \mathbf{x}, \hat{\mathbf{y}}_{\boldsymbol{\rho}_{<t}}, \rho_t)$ autoregressively (see Alg.~\ref{alg:inference}).

\subsection{Reinforced Training for Adaptive Order Prediction}\label{sec:alg}

With an RL formulation in place to solve Eq.~\ref{eq:rl}, as illustrated in Fig.~\ref{fig:method}, we use a discrete version of soft Q-learning~\cite{haarnoja2017softql} that optimizes the following entropy-regularized discounted-return objective:
\begin{equation}\label{eq:maxentrl}
    \max_\theta \mathbb{E}_{\pi_\theta} \left[ \sum_t \gamma^t (r_t + \alpha H(\pi_\theta(\cdot \mid s_t))) \right]
\end{equation}
where $\alpha \geq 0$ is the entropy coefficient.
This objective balances exploration and exploitation under a maximum entropy RL framework by requiring the policy to act as stochastically as possible while optimizing returns.
In addition, since our action space $\mathcal{A}=[M]$ is discrete and tractable, we can efficiently compute $\pi_\theta$ and $V_\theta$ from $Q_\theta$ without separately learning the policy and value function through function approximation.
To optimize this objective, we use soft Bellman update~\cite{haarnoja2017softql} implemented with the following mean squared error loss:
\begin{equation}\label{eq:sql_mse}
    \mathcal{L}_\text{SQL-MSE}(\theta) :=\mathbb{E}_{s,a,r,s'} \left[ (Q_\theta(s, a) - \bar{Q}_\theta(s,a))^2 \right]
\end{equation}
where $\bar{Q}_\theta(s,a):=r + \gamma V_\theta(s')=r+\gamma \alpha \log \sum_{a'} \exp (Q_\theta(s, a')/\alpha)$ is the Q-value target with gradients detached.
Note that we do not use target network techniques.

Alternatively, we can also use the binary cross-entropy loss as value loss:
\begin{equation}\label{eq:sql_bce}
    \mathcal{L}_\text{SQL-BCE}(\theta) :=\mathbb{E}_{s,a,r,s'} \left[ -\exp \bar{Q}_\theta(s,a) \cdot Q_\theta(s, a) - (1-\exp \bar{Q}_\theta(s,a)) \cdot \log (1- \exp Q_\theta(s, a))\right]
\end{equation}
which may deliver better robustness to outliers for $0 \leq \exp \bar{Q}_\theta(s,a) \leq 1$ and is compatible with sparse reward functions, as discussed below.

Finally, we discuss the optimization of the token predictor parameters $\psi$ and the choice of reward function.
We train the token predictor $p_\psi$ online along with the policy $\pi_\theta$ with the following (permuted) language modelling loss, echoing Eq.~\ref{eq:main_obj}:
\begin{equation}\label{eq:lm_loss}
    \mathcal{L}_\text{LM}(\psi) := \mathbb{E}_{(\mathbf{x}, \mathbf{y}) \sim D, \boldsymbol{\rho} \sim \pi_\theta} \left[ \sum_{i=1}^M -\log p_\psi(y_{\rho_i} \mid \mathbf{x}, \mathbf{y}_{\boldsymbol{\rho}_{<i}}, \rho_i) \right]
\end{equation}
Note that $p_\psi$ is trained on $\boldsymbol{\rho}$ sampled from the current policy $\pi_\theta$ to ensure that it always stays on-policy.
As $\mathcal{L}_\text{LM}$ is supervised, $p_\psi$ learns much faster than the RL-trained $\pi_\theta$, and we treat $\psi$ as always being near the optimality and approximate $I_\mathcal{V}(X,\{Y_{\rho_j}\}_{j<i} \rightarrow Y_{\rho_i})$ with the current $\log p_\psi(y_{\rho_i} \mid \mathbf{x}, \mathbf{y}_{\boldsymbol{\rho}_{<i}}, \rho_i)$.
As a result, we may set the reward function to the (negated) perplexity:
\begin{equation}\label{eq:reward_ppl}
    R(s_t, a_t)=R_\text{ppl}((\mathbf{x}, \mathbf{y}_{\boldsymbol{\rho}_{<t}}), \rho_t):=\log p_\psi(y_{\rho_t} \mid \mathbf{x}, \mathbf{y}_{\boldsymbol{\rho}_{<t}}, \rho_t)
\end{equation}
Empirically, we found the following thresholded, sparse reward with a better performance:
\begin{equation}\label{eq:reward_sparse}
    R(s_t, a_t)=R_\text{spr}((\mathbf{x}, \mathbf{y}_{\boldsymbol{\rho}_{<t}}), \rho_t):=\log \mathbb{I} [p_\psi(y_{\rho_t} \mid \mathbf{x}, \mathbf{y}_{\boldsymbol{\rho}_{<t}}, \rho_t) \geq \eta]
\end{equation}
where $\eta \in (0, 1)$ is the probability threshold.
Note, however, that the logarithm is undefined when $p_\psi(y_{\rho_i} \mid \mathbf{x}, \mathbf{y}_{\boldsymbol{\rho}_{<i}}, \rho_i) < \eta$; we use this reward in conjunction with $\mathcal{L}_\text{SQL-BCE}$, which only requires $\exp R_\text{spr}((\mathbf{x}, \mathbf{y}_{\boldsymbol{\rho}_{<t}}), \rho_t):= \mathbb{I} [p_\psi(y_{\rho_i} \mid \mathbf{x}, \mathbf{y}_{\boldsymbol{\rho}_{<i}}, \rho_i) \geq \eta]$ that is well-defined.

Training pseudocode is presented in Alg.~\ref{alg:training}.
Furthermore, since the dynamics of the MDP we defined are fully known, we can sample potentially multiple actions $\bar{\boldsymbol{\rho}}_{t, 1 \ldots K}$ at the same state $(\mathbf{x}, \mathbf{y}_{\boldsymbol{\rho}_{<t}})$ and optimize $\mathcal{L}_\text{LM}$ and $\mathcal{L}_\text{SQL}$ for all of the $K$ actions to improve learning quality.
See Sec.~\ref{sec:scaling} for details.

\subsection{Multi-stream Architecture for Order and Token Predictions}\label{sec:arch}

With the algorithm of ReCOR described above, we now instantiate the $\rho$-generator $\pi_\theta(\cdot \mid \mathbf{x}, \mathbf{y}_{\boldsymbol{\rho}_{<t}})$ and token predictor $p_\psi(\cdot \mid \mathbf{x}, \mathbf{y}_{\boldsymbol{\rho}_{<t}}, \rho_t)$.
For our experiments, we use GPT-2~\cite{radford2019gpt2} as the backbone, which could be replaced with other transformer variants.
To encode a permuted response, we add the absolute positional embedding of GPT-2 at position $\rho_i$ to token $y_{\rho_i}$, while the prompt $\mathbf{x}$ is encoded as a regular left-to-right sequence before the response tokens.
We use a full mask on the prompt $\mathbf{x}$ and a causal mask on the response $\mathbf{y}_{\boldsymbol{\rho}}$ to enable parallelized training and KV-cached inference.

This suffices to handle the sequential inputs $\mathbf{x}, \mathbf{y}_{\boldsymbol{\rho}}$.
The token predictor $p_\psi(\cdot \mid \mathbf{x}, \mathbf{y}_{\boldsymbol{\rho}_{<t}}, \rho_t)$ takes an additional scalar query position $\rho_t$ as input at every time step $t$.
Ideally, we'd like the queries at every time step not to interfere with each other; consequently, we adopt a multi-stream architecture conceptually similar to XLNet~\cite{yang2019xlnet} where a \textbf{main stream} takes $\mathbf{x}, \mathbf{y}_{\boldsymbol{\rho}}$ as input and performs self-attention as in regular transformers, and a \textbf{token query stream} takes $\bar{\boldsymbol{\rho}}$ as queries and cross-attends onto the main stream hidden states as keys and values without self-attention between queries.
This guarantees that every query would be independent of others.
We produce token predictions via a token head on the outputs of the token query stream.
For order predictions $\pi_\theta$, we can directly use the main stream outputs since order predictions do not depend on additional positions.
To further enhance expressiveness, we can optionally use an \textbf{order query stream} instead of the main stream to perform order predictions.
With $C$ learnable order query embeddings $q_{1 \ldots C}$ as inputs, we forward the order query stream for $C$ times and concatenate the outputs before projecting to Q values.
This design allows us to scale compute for better performance; see Sec.~\ref{sec:scaling} for details.

\section{Experiments}\label{sec:exp}

In this section, we aim to answer the following questions with our experiments:
1) Can ReCOR solve arithmetic problems without special data preprocessing?
2) Can ReCOR solve reasoning and planning problems adaptively without annotations?
3) Do we need adaptive orders during training or for inference only?
4) How does ReCOR compare with the state-of-the-art methods under fair inference compute settings?
5) Can the performance of ReCOR scale with more compute?

\subsection{Experiment Setup}\label{sec:exp_setup}

We chose the following datasets to validate ReCOR's performance at adaptive reasoning:
1) Arithmetic datasets, including synthetic autoregression and multiplication, which are originally generated in reverse order during data generation.
2) Puzzle datasets, including Sudoku and Zebra~\cite{shah2024causal}, which may require an adaptive, data-dependent order to solve.
We use accuracy (full response match) as the main evaluation metric and report standard deviations over 3 training seeds.

We compare with the following representative baselines:
1) Causal language models (CLM)~\cite{radford2019gpt2} that always follow a left-to-right order.
This baseline serves to demonstrate the necessity of non-left-to-right generation orders.
2) Autoregressive models supervised with ground-truth generation orders (AR-GT)~\cite{shah2024causal} that is an oracle for ReCOR, as ReCOR does not have access to the ground-truth order during both training and inference.
3) Masked diffusion models (MDM)~\cite{austin2021structured,hoogeboom2021argmax} that randomly mask the input and perform denoising, in effect trying to learn arbitrary orders.
4) Adaptive masked diffusion models (AdaMDM)~\cite{kim2025train,ye2025beyond} that use recent state-of-the-art adaptive inference methods.

\subsection{Can ReCOR solve arithmetic problems without special data preprocessing?}\label{sec:exp_arithmetic}

\begin{table}
  \caption{Performance of ReCOR and baselines on arithmetic datasets. ReCOR outperforms baselines and is competitive with the oracle.}
  \label{tab:arithmetic_main}
  \centering
  \begin{tabular}{cccccc}
    \toprule
    Task & AR-GT & ReCOR & CLM & MDM & AdaMDM \\

    \midrule
    
    ARG & $\textcolor{gray}{0.994 \pm 0.003}$ & $\mathbf{0.987 \pm 0.007}$ & $0.017 \pm 0.002$ & $0.035 \pm 0.010$ & $0.174 \pm 0.036$ \\
    MUL & $\textcolor{gray}{0.999 \pm 0.001}$ & $\mathbf{0.964 \pm 0.007}$ & $0.594 \pm 0.018$ & $0.943 \pm 0.006$ & $0.951 \pm 0.038$ \\
    \bottomrule
  \end{tabular}
\end{table}

We start with arithmetic problems generated using a fixed, non-causal order.
We employ a synthetic Autoregression (ARG) task where each response token depends on the prompt and all response tokens after it, and a multiplication (MUL) task between 20-digit and 2-digit numbers.
It has been shown that CLMs often need manually reversed training data or additional CoT annotations to solve certain arithmetic problems~\cite{lee2024teaching,shen2023positional,zhang2024reverse,mcleish2024transformers} due to the reverse dependencies between tokens brought by carry digits.
In this work, we are interested in whether ReCOR can automatically recover the correct order without manual preprocessing or additional annotations.
Consequently, we apply the reversed order only to AR-GT and keep the natural order for the rest of the approaches.

As shown in Tab.~\ref{tab:arithmetic_main}, ReCOR achieves competitive performance compared with all baselines and outperforms CLMs by a large margin, showing that the plain next-token prediction objective is not sufficient for solving these problems.
Notably, ReCOR is competitive with AR-GT, which uses the ground-truth order during both training and inference.
This demonstrates its ability to recover order in a self-supervised manner.
Furthermore, ReCOR outperforms (Ada)MDM significantly in Autoregression; we compare both methods in more detail in Sec.~\ref{sec:exp_inference_failure}.

\subsection{Can ReCOR solve reasoning and planning problems adaptively without annotations?}\label{sec:exp_puzzle}

\begin{table}
  \caption{Performance of ReCOR and baselines on puzzle datasets. ReCOR outperforms all approaches, even including the oracle AR-GT supervised with ground-truth orders.}
  \label{tab:puzzle_main}
  \centering
  \begin{tabular}{cccccc}
    \toprule
    Task & AR-GT & ReCOR & CLM & MDM & AdaMDM \\

    \midrule
    Sudoku & $\textcolor{gray}{0.8718}$ & $\mathbf{0.9017 \pm 0.0004}$ & $0.0973$ & $0.0688$ & $0.8949$ \\
    Zebra & $\textcolor{gray}{0.9117}$ & $\mathbf{0.9905 \pm 0.0021}$ & $0.8031$ & $0.769$ & $0.985$ \\
    \bottomrule
  \end{tabular}
\end{table}

In this section, we study classic logic puzzles Sudoku and Zebra~\cite{shah2024causal}.
The two datasets are examples of problems that require adaptive, data-dependent reasoning and planning, rendering them difficult for approaches that generate solutions with a fixed order.
For example, in Sudoku, we may need to fill certain cells first in order to eliminate candidates for other cells and arrive at the correct answer, as illustrated in Fig.~\ref{fig:illustration}.
The cell dependency structure varies between instances, posing a great challenge.

We show results on Sudoku and Zebra in Tab.~\ref{tab:puzzle_main}.
Values without standard deviations are reported by \cite{kim2025train}.
ReCOR is the best among all approaches, outperforming even the oracle AR-GT.
We remark that at every time step, ReCOR can generate an estimated reward for every unfilled cell, which is a denser and more diverse signal than the single next position offered by the ground truth, potentially contributing to its superior performance.
CLM and vanilla MDM perform similarly and are both worse than ReCOR and AdaMDM, showing that an adaptive generation order is indeed necessary.

\subsection{Do we need adaptive orders during training or for inference only?}
\label{sec:exp_inference_failure}

\begin{table}
  \caption{Impact of \textit{suffix unmasking} for MDMs on Autoregression. Suffix unmasking dramatically improves the performance of AdaSufMDM over AdaMDM, but still lags behind ReCOR.}
  \label{tab:sufmdm_autoreg}
  \centering
  \begin{tabular}{ccccc}
    \toprule
    ReCOR & SufMDM & AdaSufMDM & MDM & AdaMDM \\
    \midrule
    $\mathbf{0.987 \pm 0.007}$ & $0.040 \pm 0.005$ & \underline{$0.536 \pm 0.051$} & $0.035 \pm 0.010$ & $0.174 \pm 0.036$ \\
    \bottomrule
  \end{tabular}
\end{table}

We look closer into the difference between ReCOR and AdaMDM~\cite{kim2025train,ye2025beyond}, which are recent, state-of-the-art methods also focused on the order problem.
Both works showed that MDMs encounter intractable sub-problems due to random masking during training, and proposed to use adaptive decoding strategies during inference to prioritize easy tokens and circumvent these hard scenarios.
One of the core differences between ReCOR and AdaMDM is that AdaMDM is trained under random masking while ReCOR follows the recovered order during both training and inference.
Thus a problem arises: \textit{do we really need adaptive orders during training?}

We answer the question in the affirmative with the Autoregression task (Tab.~\ref{tab:arithmetic_main}).
In Autoregression, each response token is generated conditioned on \textit{all} response tokens behind it, and consequently, can only be tractably predicted when all of these tokens are present in the context.
Similar properties hold for many long-horizon reasoning problems with multiple steps and dense dependencies between the steps.
However, under random masking, the probability of such an event grows exponentially smaller with the length of the context, and MDMs rarely see any long, complete contexts during training.
As a result, \textbf{MDMs are bad at the corresponding sub-problems, even though they are supposedly easy and tractable}, leading to a bad overall performance as evidenced in Tab.~\ref{tab:arithmetic_main}.
In contrast, \textbf{ReCOR automatically recognizes and frequently visits these good sub-problems during training}, making sure that the on-policy token predictor $p_\psi$ can reliably solve them.

To further support this analysis, we propose (Ada)SufMDM, a variant of (Ada)MDM that \textit{unmasks} a suffix of length $l \sim \mathcal{U}_{[M]} - 1$ during training.
This operation dramatically boosts the probability of each "correct" sub-problem from $O(c^M)$ to $\Omega(M^{-1})$.
Note that this fix applies only to ARG, where the correct order is known to be reversed; applying the operation in general would require ground-truth order annotations.
The updated results are shown in Tab.~\ref{tab:sufmdm_autoreg}.
SufAdaMDM obtains a large performance improvement and still outperforms SufMDM, demonstrating that correct orders are important during both training and inference.
However, there remains a gap between AdaSufMDM and ReCOR, which is likely due to the additional intractable sub-problems brought by random masking, as argued \cite{kim2025train}.
Our analysis complements their findings in that \textbf{random masking leads to not only many intractable sub-problems, but also a dearth of tractable ones}.

\subsection{How does ReCOR compare with the state-of-the-art methods under fair inference compute settings?}\label{sec:exp_computation}

\begin{table}
  \caption{Performance of ReCOR and Ada(Suf)MDM under different inference compute settings. MDM underperforms ReCOR even with 10x compute, and is much worse with the same compute. We use AdaMDM on MUL and AdaSufMDM on ARG, as AdaMDM fails on ARG.}
  \label{tab:inference_compute}
  \centering
  \begin{tabular}{cccc}
    \toprule
    Method & ReCOR & \multicolumn{2}{c}{Ada(Suf)MDM} \\

    \midrule
    FLOPs & 1x & 1x($T=2$) & 10x($T=20$) \\
    \midrule
    ARG & $\mathbf{0.987 \pm 0.007}$ & $0.028 \pm 0.009$ & $0.536 \pm 0.051$ \\
    MUL & $\mathbf{0.964 \pm 0.007}$ & $0.875 \pm 0.043$ & $0.951 \pm 0.038$ \\
    \bottomrule
  \end{tabular}
\end{table}

We show another advantage of our autoregressive design by comparing ReCOR with MDMs under different inference compute settings.
The standard MDM inference setting used in our primary experiments employs a large number of diffusion steps; consequently, many transformer forward passes are needed during inference.
We note that for inherently serial problems where tokens need to be generated one-by-one, MDMs require diffusion steps to the order of $\Omega(M)$, bringing the overall time complexity to $\Omega(M^3)$.
In comparison, since ReCOR is autoregressive and uses a causal mask for the response, we can use KV-caching during inference, reducing the total number of complete transformer forward passes to $2+C$ where $C$ is the number of order stream queries and a constant with respect to the context length, maintaining an $O(M^2)$ time complexity.

Empirically, we infer MDMs on the arithmetic datasets under alternative inference compute settings in Tab.~\ref{tab:inference_compute}.
ReCOR uses $C=0$ for these datasets, leading to an inference FLOPs equivalent to 2 complete transformer forward passes.
When restricting the FLOPs of MDMs to the same level, the number of diffusion steps is limited to $T=2$, dramatically reducing performance.

\subsection{Can the performance of ReCOR scale with more compute?}\label{sec:scaling}

\begin{figure}[t]
  \centering
  \begin{subfigure}{0.49\textwidth}
      \includegraphics[width=\linewidth]{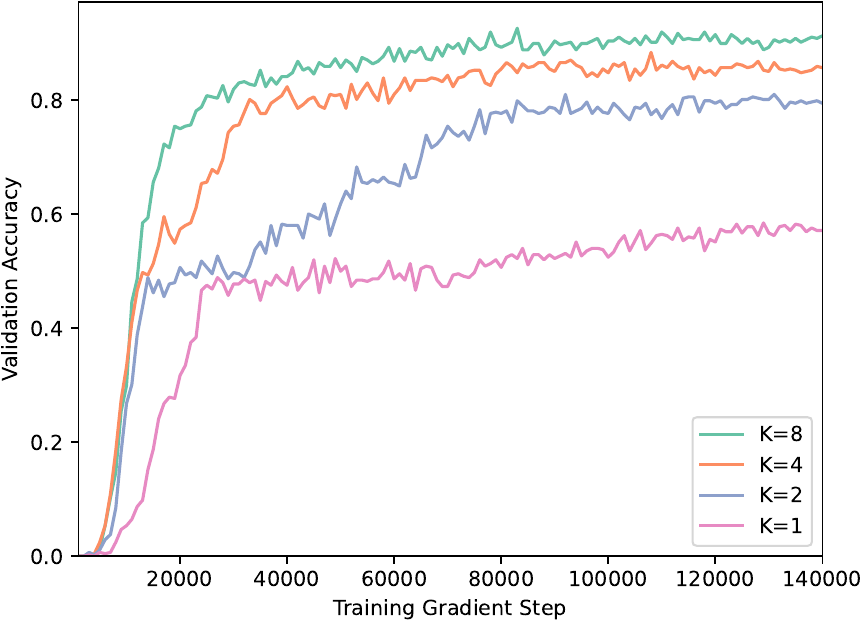}
      \subcaption{Token query scaling with $C=8$.}
      \label{fig:scaling_token}
  \end{subfigure}
  \begin{subfigure}{0.49\textwidth}
      \includegraphics[width=\linewidth]{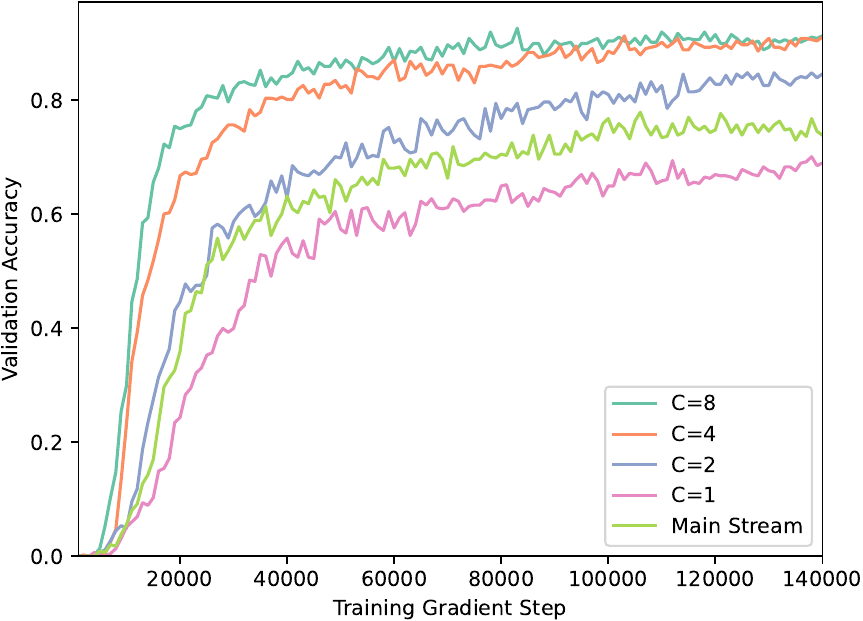}
      \subcaption{Order query scaling with $K=8$.}
      \label{fig:scaling_order}
  \end{subfigure}
  
  \caption{ReCOR's performance when scaling the number of token queries $K$ (a) and order queries $C$ (b). Main Stream in (b) denotes using the main stream outputs without a separate order query stream. ReCOR can improve its performance with more computation during training and inference.}
  \label{fig:scaling}
\end{figure}

In this section, we demonstrate scaling properties arising from the designs of ReCOR.
As described in Sec.~\ref{sec:method}, in contrast to the standard RL setting, ReCOR can take $K>1$ actions at the same state and learn from all of these actions, since the MDP we defined can be easily simulated perfectly.
Furthermore, we can use $C$ queries to the order query stream to obtain a more expressive Q function.
These designs allow us to scale compute to improve performance even with the underlying backbone parameter count fixed.

To demonstrate the scaling properties, we perform ablation experiments on Sudoku that vary one of $K$ or $C$ from the primary setting $K=C=8$.
We show training curves of validation accuracy in Fig.~\ref{fig:scaling}.
It can be seen that in both groups of experiments, performance improves monotonically with the varying parameter.
As a special case, directly using the main stream outputs to compute Q values (Main Stream in Fig.~\ref{fig:scaling_order}) achieves decent performance between $C=1$ and $2$ without using separate order queries.
This makes it a good, lightweight choice that we adopt for the arithmetic datasets.

\section{Conclusion and Limitations}\label{sec:conclusion}

In this paper, we introduce Reinforced Context Order Recovery (ReCOR), a reinforcement-learning-based algorithm for automatically recovering the correct generation order from textual data without annotations.
While modern text generation models are predominantly causal, they have been shown to fail in adaptive reasoning problems due to the presence of intractable tokens.
We characterize this phenomenon with the framework of $\mathcal{V}$-information and propose to optimize the corresponding objective to learn a suitable data-dependent generation order.
Subsequently, we operationalize this objective by introducing ReCOR with reinforcement learning setups, recovering unobserved generation order from purely textual data in a self-supervised manner.
Empirically, ReCOR outperforms strong baselines on various datasets, including recent adaptive inference approaches using masked diffusion models and oracle models supervised with the ground-truth order.

There are certain limitations and future works for ReCOR.
We primarily run experiments on reasoning and planning-related datasets, echoing the setups of our baselines.
Going forward, we are excited to further scale up ReCOR to more diverse problems and datasets with more compute available.
The RL-based formulation also opens up a vast space to integrate with other RL techniques and further improve the performance of ReCOR, which we did not exhaust due to limited resources.
Furthermore, there could be more unobserved variables driving the generation of textual data beyond the generation order.
The recovery of these variables presents exciting opportunities for future explorations.

\bibliographystyle{unsrt}
\bibliography{main}
\clearpage
\appendix

\section{Broader Impacts}\label{sec:broader_impacts}

ReCOR aims to enhance the adaptive reasoning and planning capabilities of machine learning models, with potential applications in many fields, e.g. code writing, assisting scientific discoveries, and general LLM agents etc.
Embodied agents with real-world tasks also require these capabilities.
We regard ReCOR as mostly foundational research while noting that advancements in such capabilities have both positive and negative potential consequences if subject to misuse.

\section{Pseudocode of Training and Inference Algorithms for ReCOR}

\begin{algorithm}
\caption{Training of ReCOR.}\label{alg:training}
\begin{algorithmic}
\Require Dataset $D$
\Ensure Trained parameters $\theta, \psi$
\State Initialize $\theta, \psi$
\While{Not converged}
\State Sample mini-batch $B=\{(\mathbf{x}, \mathbf{y})\} \sim D$
\For{$t =1\ldots M$} \Comment{Rollout $\pi_\theta$ on $B$}
    \State Sample $\rho_t, \bar{\boldsymbol{\rho}}_{t, [K]} \sim \pi_\theta(\cdot \mid \mathbf{x}, \mathbf{y}_{\boldsymbol{\rho}_{<t}})$
\EndFor
\State Compute rewards $\mathbf{r}$ using $\psi$ on $B,\boldsymbol{\rho}, \bar{\boldsymbol{\rho}}$
\State Update $\psi$ with $\mathcal{L}_\text{LM}$ and $\theta$ with $\mathcal{L}_\text{SQL}$ on $B,\boldsymbol{\rho}, \bar{\boldsymbol{\rho}},\mathbf{r}$
\EndWhile
\end{algorithmic}
\end{algorithm}

\begin{algorithm}
\caption{Inference of ReCOR.}\label{alg:inference}
\begin{algorithmic}
\Require Evaluation prompt $\mathbf{x}$, parameters $\theta^*, \psi^*$
\Ensure Generated response $\hat{\mathbf{y}}$
\For{$t =1\ldots M$}
    \State Sample $\rho_t \sim \pi_{\theta^*}(\cdot \mid \mathbf{x}, \hat{\mathbf{y}}_{\boldsymbol{\rho}_{<t}})$
    \State Sample $\hat{y}_{\rho_t} \sim p_{\psi^*}(\cdot \mid \mathbf{x}, \hat{\mathbf{y}}_{\boldsymbol{\rho}_{<t}}, \rho_t)$
\EndFor
\end{algorithmic}
\end{algorithm}

\section{Generation Examples}

Here we show generation examples of ReCOR and AdaMDM on Autoregression.
The prompt is \verb|64610246434563440135| while the correct response is \verb|15443515654216654535|.
Grey \colorbox{gray!30}{*} denotes unfilled positions, \colorbox{green!30}{green} denotes correct tokens, and \colorbox{red!30}{red} ones are incorrect tokens.

As observed below, ReCOR (left) recovers the reversed generation order and generates the correct tokens, while AdaMDM (right) encounters difficulties and incorrectly generates later tokens.
Note that even though AdaMDM uses adaptive inference techniques, it is confused about which token can actually be predicted correctly, and hallucinates wrong tokens.
We analyze and explain this phenomenon in the experiment section in the main text.

\begin{minipage}{0.48\textwidth}
\colorbox{gray!30}{*}\colorbox{gray!30}{*}\colorbox{gray!30}{*}\colorbox{gray!30}{*}\colorbox{gray!30}{*}\colorbox{gray!30}{*}\colorbox{gray!30}{*}\colorbox{gray!30}{*}\colorbox{gray!30}{*}\colorbox{gray!30}{*}\colorbox{gray!30}{*}\colorbox{gray!30}{*}\colorbox{gray!30}{*}\colorbox{gray!30}{*}\colorbox{gray!30}{*}\colorbox{gray!30}{*}\colorbox{gray!30}{*}\colorbox{gray!30}{*}\colorbox{gray!30}{*}\colorbox{green!30}{5}
\colorbox{gray!30}{*}\colorbox{gray!30}{*}\colorbox{gray!30}{*}\colorbox{gray!30}{*}\colorbox{gray!30}{*}\colorbox{gray!30}{*}\colorbox{gray!30}{*}\colorbox{gray!30}{*}\colorbox{gray!30}{*}\colorbox{gray!30}{*}\colorbox{gray!30}{*}\colorbox{gray!30}{*}\colorbox{gray!30}{*}\colorbox{gray!30}{*}\colorbox{gray!30}{*}\colorbox{gray!30}{*}\colorbox{gray!30}{*}\colorbox{gray!30}{*}\colorbox{green!30}{3}\colorbox{green!30}{5}
\colorbox{gray!30}{*}\colorbox{gray!30}{*}\colorbox{gray!30}{*}\colorbox{gray!30}{*}\colorbox{gray!30}{*}\colorbox{gray!30}{*}\colorbox{gray!30}{*}\colorbox{gray!30}{*}\colorbox{gray!30}{*}\colorbox{gray!30}{*}\colorbox{gray!30}{*}\colorbox{gray!30}{*}\colorbox{gray!30}{*}\colorbox{gray!30}{*}\colorbox{gray!30}{*}\colorbox{gray!30}{*}\colorbox{gray!30}{*}\colorbox{green!30}{5}\colorbox{green!30}{3}\colorbox{green!30}{5}
\colorbox{gray!30}{*}\colorbox{gray!30}{*}\colorbox{gray!30}{*}\colorbox{gray!30}{*}\colorbox{gray!30}{*}\colorbox{gray!30}{*}\colorbox{gray!30}{*}\colorbox{gray!30}{*}\colorbox{gray!30}{*}\colorbox{gray!30}{*}\colorbox{gray!30}{*}\colorbox{gray!30}{*}\colorbox{gray!30}{*}\colorbox{gray!30}{*}\colorbox{gray!30}{*}\colorbox{gray!30}{*}\colorbox{green!30}{4}\colorbox{green!30}{5}\colorbox{green!30}{3}\colorbox{green!30}{5}
\colorbox{gray!30}{*}\colorbox{gray!30}{*}\colorbox{gray!30}{*}\colorbox{gray!30}{*}\colorbox{gray!30}{*}\colorbox{gray!30}{*}\colorbox{gray!30}{*}\colorbox{gray!30}{*}\colorbox{gray!30}{*}\colorbox{gray!30}{*}\colorbox{gray!30}{*}\colorbox{gray!30}{*}\colorbox{gray!30}{*}\colorbox{gray!30}{*}\colorbox{gray!30}{*}\colorbox{green!30}{5}\colorbox{green!30}{4}\colorbox{green!30}{5}\colorbox{green!30}{3}\colorbox{green!30}{5}
\colorbox{gray!30}{*}\colorbox{gray!30}{*}\colorbox{gray!30}{*}\colorbox{gray!30}{*}\colorbox{gray!30}{*}\colorbox{gray!30}{*}\colorbox{gray!30}{*}\colorbox{gray!30}{*}\colorbox{gray!30}{*}\colorbox{gray!30}{*}\colorbox{gray!30}{*}\colorbox{gray!30}{*}\colorbox{gray!30}{*}\colorbox{gray!30}{*}\colorbox{green!30}{6}\colorbox{green!30}{5}\colorbox{green!30}{4}\colorbox{green!30}{5}\colorbox{green!30}{3}\colorbox{green!30}{5}
\colorbox{gray!30}{*}\colorbox{gray!30}{*}\colorbox{gray!30}{*}\colorbox{gray!30}{*}\colorbox{gray!30}{*}\colorbox{gray!30}{*}\colorbox{gray!30}{*}\colorbox{gray!30}{*}\colorbox{gray!30}{*}\colorbox{gray!30}{*}\colorbox{gray!30}{*}\colorbox{gray!30}{*}\colorbox{gray!30}{*}\colorbox{green!30}{6}\colorbox{green!30}{6}\colorbox{green!30}{5}\colorbox{green!30}{4}\colorbox{green!30}{5}\colorbox{green!30}{3}\colorbox{green!30}{5}
\colorbox{gray!30}{*}\colorbox{gray!30}{*}\colorbox{gray!30}{*}\colorbox{gray!30}{*}\colorbox{gray!30}{*}\colorbox{gray!30}{*}\colorbox{gray!30}{*}\colorbox{gray!30}{*}\colorbox{gray!30}{*}\colorbox{gray!30}{*}\colorbox{gray!30}{*}\colorbox{gray!30}{*}\colorbox{green!30}{1}\colorbox{green!30}{6}\colorbox{green!30}{6}\colorbox{green!30}{5}\colorbox{green!30}{4}\colorbox{green!30}{5}\colorbox{green!30}{3}\colorbox{green!30}{5}
\colorbox{gray!30}{*}\colorbox{gray!30}{*}\colorbox{gray!30}{*}\colorbox{gray!30}{*}\colorbox{gray!30}{*}\colorbox{gray!30}{*}\colorbox{gray!30}{*}\colorbox{gray!30}{*}\colorbox{gray!30}{*}\colorbox{gray!30}{*}\colorbox{gray!30}{*}\colorbox{green!30}{2}\colorbox{green!30}{1}\colorbox{green!30}{6}\colorbox{green!30}{6}\colorbox{green!30}{5}\colorbox{green!30}{4}\colorbox{green!30}{5}\colorbox{green!30}{3}\colorbox{green!30}{5}
\colorbox{gray!30}{*}\colorbox{gray!30}{*}\colorbox{gray!30}{*}\colorbox{gray!30}{*}\colorbox{gray!30}{*}\colorbox{gray!30}{*}\colorbox{gray!30}{*}\colorbox{gray!30}{*}\colorbox{gray!30}{*}\colorbox{gray!30}{*}\colorbox{green!30}{4}\colorbox{green!30}{2}\colorbox{green!30}{1}\colorbox{green!30}{6}\colorbox{green!30}{6}\colorbox{green!30}{5}\colorbox{green!30}{4}\colorbox{green!30}{5}\colorbox{green!30}{3}\colorbox{green!30}{5}
\colorbox{gray!30}{*}\colorbox{gray!30}{*}\colorbox{gray!30}{*}\colorbox{gray!30}{*}\colorbox{gray!30}{*}\colorbox{gray!30}{*}\colorbox{gray!30}{*}\colorbox{gray!30}{*}\colorbox{gray!30}{*}\colorbox{green!30}{5}\colorbox{green!30}{4}\colorbox{green!30}{2}\colorbox{green!30}{1}\colorbox{green!30}{6}\colorbox{green!30}{6}\colorbox{green!30}{5}\colorbox{green!30}{4}\colorbox{green!30}{5}\colorbox{green!30}{3}\colorbox{green!30}{5}
\colorbox{gray!30}{*}\colorbox{gray!30}{*}\colorbox{gray!30}{*}\colorbox{gray!30}{*}\colorbox{gray!30}{*}\colorbox{gray!30}{*}\colorbox{gray!30}{*}\colorbox{gray!30}{*}\colorbox{green!30}{6}\colorbox{green!30}{5}\colorbox{green!30}{4}\colorbox{green!30}{2}\colorbox{green!30}{1}\colorbox{green!30}{6}\colorbox{green!30}{6}\colorbox{green!30}{5}\colorbox{green!30}{4}\colorbox{green!30}{5}\colorbox{green!30}{3}\colorbox{green!30}{5}
\colorbox{gray!30}{*}\colorbox{gray!30}{*}\colorbox{gray!30}{*}\colorbox{gray!30}{*}\colorbox{gray!30}{*}\colorbox{gray!30}{*}\colorbox{gray!30}{*}\colorbox{green!30}{5}\colorbox{green!30}{6}\colorbox{green!30}{5}\colorbox{green!30}{4}\colorbox{green!30}{2}\colorbox{green!30}{1}\colorbox{green!30}{6}\colorbox{green!30}{6}\colorbox{green!30}{5}\colorbox{green!30}{4}\colorbox{green!30}{5}\colorbox{green!30}{3}\colorbox{green!30}{5}
\colorbox{gray!30}{*}\colorbox{gray!30}{*}\colorbox{gray!30}{*}\colorbox{gray!30}{*}\colorbox{gray!30}{*}\colorbox{gray!30}{*}\colorbox{green!30}{1}\colorbox{green!30}{5}\colorbox{green!30}{6}\colorbox{green!30}{5}\colorbox{green!30}{4}\colorbox{green!30}{2}\colorbox{green!30}{1}\colorbox{green!30}{6}\colorbox{green!30}{6}\colorbox{green!30}{5}\colorbox{green!30}{4}\colorbox{green!30}{5}\colorbox{green!30}{3}\colorbox{green!30}{5}
\colorbox{gray!30}{*}\colorbox{gray!30}{*}\colorbox{gray!30}{*}\colorbox{gray!30}{*}\colorbox{gray!30}{*}\colorbox{green!30}{5}\colorbox{green!30}{1}\colorbox{green!30}{5}\colorbox{green!30}{6}\colorbox{green!30}{5}\colorbox{green!30}{4}\colorbox{green!30}{2}\colorbox{green!30}{1}\colorbox{green!30}{6}\colorbox{green!30}{6}\colorbox{green!30}{5}\colorbox{green!30}{4}\colorbox{green!30}{5}\colorbox{green!30}{3}\colorbox{green!30}{5}
\colorbox{gray!30}{*}\colorbox{gray!30}{*}\colorbox{gray!30}{*}\colorbox{gray!30}{*}\colorbox{green!30}{3}\colorbox{green!30}{5}\colorbox{green!30}{1}\colorbox{green!30}{5}\colorbox{green!30}{6}\colorbox{green!30}{5}\colorbox{green!30}{4}\colorbox{green!30}{2}\colorbox{green!30}{1}\colorbox{green!30}{6}\colorbox{green!30}{6}\colorbox{green!30}{5}\colorbox{green!30}{4}\colorbox{green!30}{5}\colorbox{green!30}{3}\colorbox{green!30}{5}
\colorbox{gray!30}{*}\colorbox{gray!30}{*}\colorbox{green!30}{4}\colorbox{gray!30}{*}\colorbox{green!30}{3}\colorbox{green!30}{5}\colorbox{green!30}{1}\colorbox{green!30}{5}\colorbox{green!30}{6}\colorbox{green!30}{5}\colorbox{green!30}{4}\colorbox{green!30}{2}\colorbox{green!30}{1}\colorbox{green!30}{6}\colorbox{green!30}{6}\colorbox{green!30}{5}\colorbox{green!30}{4}\colorbox{green!30}{5}\colorbox{green!30}{3}\colorbox{green!30}{5}
\colorbox{gray!30}{*}\colorbox{gray!30}{*}\colorbox{green!30}{4}\colorbox{green!30}{4}\colorbox{green!30}{3}\colorbox{green!30}{5}\colorbox{green!30}{1}\colorbox{green!30}{5}\colorbox{green!30}{6}\colorbox{green!30}{5}\colorbox{green!30}{4}\colorbox{green!30}{2}\colorbox{green!30}{1}\colorbox{green!30}{6}\colorbox{green!30}{6}\colorbox{green!30}{5}\colorbox{green!30}{4}\colorbox{green!30}{5}\colorbox{green!30}{3}\colorbox{green!30}{5}
\colorbox{gray!30}{*}\colorbox{green!30}{5}\colorbox{green!30}{4}\colorbox{green!30}{4}\colorbox{green!30}{3}\colorbox{green!30}{5}\colorbox{green!30}{1}\colorbox{green!30}{5}\colorbox{green!30}{6}\colorbox{green!30}{5}\colorbox{green!30}{4}\colorbox{green!30}{2}\colorbox{green!30}{1}\colorbox{green!30}{6}\colorbox{green!30}{6}\colorbox{green!30}{5}\colorbox{green!30}{4}\colorbox{green!30}{5}\colorbox{green!30}{3}\colorbox{green!30}{5}
\colorbox{green!30}{1}\colorbox{green!30}{5}\colorbox{green!30}{4}\colorbox{green!30}{4}\colorbox{green!30}{3}\colorbox{green!30}{5}\colorbox{green!30}{1}\colorbox{green!30}{5}\colorbox{green!30}{6}\colorbox{green!30}{5}\colorbox{green!30}{4}\colorbox{green!30}{2}\colorbox{green!30}{1}\colorbox{green!30}{6}\colorbox{green!30}{6}\colorbox{green!30}{5}\colorbox{green!30}{4}\colorbox{green!30}{5}\colorbox{green!30}{3}\colorbox{green!30}{5}

\end{minipage}
\begin{minipage}{0.48\textwidth}
\colorbox{gray!30}{*}\colorbox{gray!30}{*}\colorbox{gray!30}{*}\colorbox{gray!30}{*}\colorbox{gray!30}{*}\colorbox{gray!30}{*}\colorbox{gray!30}{*}\colorbox{gray!30}{*}\colorbox{gray!30}{*}\colorbox{gray!30}{*}\colorbox{gray!30}{*}\colorbox{gray!30}{*}\colorbox{gray!30}{*}\colorbox{green!30}{6}\colorbox{gray!30}{*}\colorbox{gray!30}{*}\colorbox{gray!30}{*}\colorbox{gray!30}{*}\colorbox{gray!30}{*}\colorbox{gray!30}{*}
\colorbox{gray!30}{*}\colorbox{gray!30}{*}\colorbox{gray!30}{*}\colorbox{gray!30}{*}\colorbox{gray!30}{*}\colorbox{gray!30}{*}\colorbox{gray!30}{*}\colorbox{gray!30}{*}\colorbox{gray!30}{*}\colorbox{gray!30}{*}\colorbox{gray!30}{*}\colorbox{green!30}{2}\colorbox{gray!30}{*}\colorbox{gray!30}{*}\colorbox{gray!30}{*}\colorbox{gray!30}{*}\colorbox{gray!30}{*}\colorbox{green!30}{5}\colorbox{gray!30}{*}\colorbox{gray!30}{*}
\colorbox{gray!30}{*}\colorbox{gray!30}{*}\colorbox{gray!30}{*}\colorbox{gray!30}{*}\colorbox{gray!30}{*}\colorbox{gray!30}{*}\colorbox{gray!30}{*}\colorbox{gray!30}{*}\colorbox{gray!30}{*}\colorbox{gray!30}{*}\colorbox{gray!30}{*}\colorbox{green!30}{2}\colorbox{gray!30}{*}\colorbox{gray!30}{*}\colorbox{gray!30}{*}\colorbox{gray!30}{*}\colorbox{green!30}{4}\colorbox{green!30}{5}\colorbox{green!30}{3}\colorbox{gray!30}{*}
\colorbox{gray!30}{*}\colorbox{gray!30}{*}\colorbox{gray!30}{*}\colorbox{gray!30}{*}\colorbox{gray!30}{*}\colorbox{gray!30}{*}\colorbox{gray!30}{*}\colorbox{gray!30}{*}\colorbox{gray!30}{*}\colorbox{gray!30}{*}\colorbox{gray!30}{*}\colorbox{gray!30}{*}\colorbox{gray!30}{*}\colorbox{green!30}{6}\colorbox{gray!30}{*}\colorbox{green!30}{5}\colorbox{green!30}{4}\colorbox{gray!30}{*}\colorbox{green!30}{3}\colorbox{gray!30}{*}
\colorbox{gray!30}{*}\colorbox{gray!30}{*}\colorbox{gray!30}{*}\colorbox{gray!30}{*}\colorbox{gray!30}{*}\colorbox{gray!30}{*}\colorbox{gray!30}{*}\colorbox{gray!30}{*}\colorbox{gray!30}{*}\colorbox{gray!30}{*}\colorbox{gray!30}{*}\colorbox{gray!30}{*}\colorbox{green!30}{1}\colorbox{gray!30}{*}\colorbox{gray!30}{*}\colorbox{green!30}{5}\colorbox{green!30}{4}\colorbox{green!30}{5}\colorbox{green!30}{3}\colorbox{gray!30}{*}
\colorbox{gray!30}{*}\colorbox{gray!30}{*}\colorbox{gray!30}{*}\colorbox{gray!30}{*}\colorbox{gray!30}{*}\colorbox{gray!30}{*}\colorbox{gray!30}{*}\colorbox{gray!30}{*}\colorbox{gray!30}{*}\colorbox{green!30}{5}\colorbox{green!30}{4}\colorbox{gray!30}{*}\colorbox{green!30}{1}\colorbox{green!30}{6}\colorbox{green!30}{6}\colorbox{gray!30}{*}\colorbox{green!30}{4}\colorbox{gray!30}{*}\colorbox{gray!30}{*}\colorbox{gray!30}{*}
\colorbox{gray!30}{*}\colorbox{gray!30}{*}\colorbox{gray!30}{*}\colorbox{gray!30}{*}\colorbox{gray!30}{*}\colorbox{gray!30}{*}\colorbox{gray!30}{*}\colorbox{green!30}{5}\colorbox{gray!30}{*}\colorbox{gray!30}{*}\colorbox{gray!30}{*}\colorbox{green!30}{2}\colorbox{green!30}{1}\colorbox{green!30}{6}\colorbox{gray!30}{*}\colorbox{green!30}{5}\colorbox{gray!30}{*}\colorbox{green!30}{5}\colorbox{green!30}{3}\colorbox{green!30}{5}
\colorbox{gray!30}{*}\colorbox{gray!30}{*}\colorbox{gray!30}{*}\colorbox{gray!30}{*}\colorbox{gray!30}{*}\colorbox{gray!30}{*}\colorbox{green!30}{1}\colorbox{gray!30}{*}\colorbox{gray!30}{*}\colorbox{green!30}{5}\colorbox{green!30}{4}\colorbox{gray!30}{*}\colorbox{gray!30}{*}\colorbox{gray!30}{*}\colorbox{gray!30}{*}\colorbox{green!30}{5}\colorbox{green!30}{4}\colorbox{green!30}{5}\colorbox{green!30}{3}\colorbox{green!30}{5}
\colorbox{gray!30}{*}\colorbox{gray!30}{*}\colorbox{gray!30}{*}\colorbox{gray!30}{*}\colorbox{gray!30}{*}\colorbox{gray!30}{*}\colorbox{green!30}{1}\colorbox{gray!30}{*}\colorbox{gray!30}{*}\colorbox{gray!30}{*}\colorbox{green!30}{4}\colorbox{green!30}{2}\colorbox{gray!30}{*}\colorbox{green!30}{6}\colorbox{green!30}{6}\colorbox{green!30}{5}\colorbox{green!30}{4}\colorbox{gray!30}{*}\colorbox{green!30}{3}\colorbox{green!30}{5}
\colorbox{gray!30}{*}\colorbox{gray!30}{*}\colorbox{gray!30}{*}\colorbox{gray!30}{*}\colorbox{gray!30}{*}\colorbox{gray!30}{*}\colorbox{green!30}{1}\colorbox{gray!30}{*}\colorbox{gray!30}{*}\colorbox{green!30}{5}\colorbox{green!30}{4}\colorbox{gray!30}{*}\colorbox{green!30}{1}\colorbox{green!30}{6}\colorbox{gray!30}{*}\colorbox{green!30}{5}\colorbox{green!30}{4}\colorbox{green!30}{5}\colorbox{green!30}{3}\colorbox{green!30}{5}
\colorbox{gray!30}{*}\colorbox{gray!30}{*}\colorbox{gray!30}{*}\colorbox{gray!30}{*}\colorbox{gray!30}{*}\colorbox{gray!30}{*}\colorbox{green!30}{1}\colorbox{green!30}{5}\colorbox{green!30}{6}\colorbox{green!30}{5}\colorbox{gray!30}{*}\colorbox{green!30}{2}\colorbox{green!30}{1}\colorbox{gray!30}{*}\colorbox{green!30}{6}\colorbox{green!30}{5}\colorbox{green!30}{4}\colorbox{green!30}{5}\colorbox{green!30}{3}\colorbox{gray!30}{*}
\colorbox{gray!30}{*}\colorbox{gray!30}{*}\colorbox{gray!30}{*}\colorbox{gray!30}{*}\colorbox{gray!30}{*}\colorbox{gray!30}{*}\colorbox{green!30}{1}\colorbox{green!30}{5}\colorbox{green!30}{6}\colorbox{green!30}{5}\colorbox{green!30}{4}\colorbox{green!30}{2}\colorbox{green!30}{1}\colorbox{green!30}{6}\colorbox{green!30}{6}\colorbox{gray!30}{*}\colorbox{gray!30}{*}\colorbox{green!30}{5}\colorbox{green!30}{3}\colorbox{green!30}{5}
\colorbox{gray!30}{*}\colorbox{gray!30}{*}\colorbox{gray!30}{*}\colorbox{gray!30}{*}\colorbox{gray!30}{*}\colorbox{gray!30}{*}\colorbox{green!30}{1}\colorbox{green!30}{5}\colorbox{green!30}{6}\colorbox{green!30}{5}\colorbox{green!30}{4}\colorbox{green!30}{2}\colorbox{green!30}{1}\colorbox{green!30}{6}\colorbox{green!30}{6}\colorbox{green!30}{5}\colorbox{green!30}{4}\colorbox{green!30}{5}\colorbox{green!30}{3}\colorbox{gray!30}{*}
\colorbox{gray!30}{*}\colorbox{gray!30}{*}\colorbox{gray!30}{*}\colorbox{gray!30}{*}\colorbox{gray!30}{*}\colorbox{gray!30}{*}\colorbox{green!30}{1}\colorbox{green!30}{5}\colorbox{green!30}{6}\colorbox{green!30}{5}\colorbox{green!30}{4}\colorbox{green!30}{2}\colorbox{green!30}{1}\colorbox{green!30}{6}\colorbox{green!30}{6}\colorbox{green!30}{5}\colorbox{green!30}{4}\colorbox{green!30}{5}\colorbox{green!30}{3}\colorbox{green!30}{5}
\colorbox{gray!30}{*}\colorbox{red!30}{1}\colorbox{gray!30}{*}\colorbox{gray!30}{*}\colorbox{gray!30}{*}\colorbox{gray!30}{*}\colorbox{green!30}{1}\colorbox{green!30}{5}\colorbox{green!30}{6}\colorbox{green!30}{5}\colorbox{green!30}{4}\colorbox{green!30}{2}\colorbox{green!30}{1}\colorbox{green!30}{6}\colorbox{green!30}{6}\colorbox{green!30}{5}\colorbox{green!30}{4}\colorbox{green!30}{5}\colorbox{green!30}{3}\colorbox{green!30}{5}
\colorbox{red!30}{4}\colorbox{red!30}{1}\colorbox{gray!30}{*}\colorbox{gray!30}{*}\colorbox{gray!30}{*}\colorbox{gray!30}{*}\colorbox{green!30}{1}\colorbox{green!30}{5}\colorbox{green!30}{6}\colorbox{green!30}{5}\colorbox{green!30}{4}\colorbox{green!30}{2}\colorbox{green!30}{1}\colorbox{green!30}{6}\colorbox{green!30}{6}\colorbox{green!30}{5}\colorbox{green!30}{4}\colorbox{green!30}{5}\colorbox{green!30}{3}\colorbox{green!30}{5}
\colorbox{red!30}{4}\colorbox{red!30}{1}\colorbox{red!30}{1}\colorbox{gray!30}{*}\colorbox{gray!30}{*}\colorbox{gray!30}{*}\colorbox{green!30}{1}\colorbox{green!30}{5}\colorbox{green!30}{6}\colorbox{green!30}{5}\colorbox{green!30}{4}\colorbox{green!30}{2}\colorbox{green!30}{1}\colorbox{green!30}{6}\colorbox{green!30}{6}\colorbox{green!30}{5}\colorbox{green!30}{4}\colorbox{green!30}{5}\colorbox{green!30}{3}\colorbox{green!30}{5}
\colorbox{red!30}{4}\colorbox{red!30}{1}\colorbox{gray!30}{*}\colorbox{red!30}{5}\colorbox{gray!30}{*}\colorbox{green!30}{5}\colorbox{green!30}{1}\colorbox{green!30}{5}\colorbox{green!30}{6}\colorbox{green!30}{5}\colorbox{green!30}{4}\colorbox{green!30}{2}\colorbox{green!30}{1}\colorbox{green!30}{6}\colorbox{green!30}{6}\colorbox{green!30}{5}\colorbox{green!30}{4}\colorbox{green!30}{5}\colorbox{green!30}{3}\colorbox{green!30}{5}
\colorbox{red!30}{4}\colorbox{red!30}{1}\colorbox{red!30}{1}\colorbox{red!30}{5}\colorbox{gray!30}{*}\colorbox{green!30}{5}\colorbox{green!30}{1}\colorbox{green!30}{5}\colorbox{green!30}{6}\colorbox{green!30}{5}\colorbox{green!30}{4}\colorbox{green!30}{2}\colorbox{green!30}{1}\colorbox{green!30}{6}\colorbox{green!30}{6}\colorbox{green!30}{5}\colorbox{green!30}{4}\colorbox{green!30}{5}\colorbox{green!30}{3}\colorbox{green!30}{5}
\colorbox{red!30}{4}\colorbox{red!30}{1}\colorbox{red!30}{1}\colorbox{red!30}{5}\colorbox{red!30}{4}\colorbox{green!30}{5}\colorbox{green!30}{1}\colorbox{green!30}{5}\colorbox{green!30}{6}\colorbox{green!30}{5}\colorbox{green!30}{4}\colorbox{green!30}{2}\colorbox{green!30}{1}\colorbox{green!30}{6}\colorbox{green!30}{6}\colorbox{green!30}{5}\colorbox{green!30}{4}\colorbox{green!30}{5}\colorbox{green!30}{3}\colorbox{green!30}{5}

\end{minipage}

\section{Experiment Details}

\subsection{Evaluation Datasets}

Here we describe details about the datasets used in our experiments.
For all datasets, we randomly split a small validation set of size 448 from the training set for validation purposes.
\subsubsection{Synthetic Autoregression}
We generate a synthetic autoregression dataset (ARG) via the following procedure:
For length $L$ and prime modulus $p$, we first generate the prompt $x_i \sim \mathcal{U}_{[p]}, 1 \leq i \leq L$, the generate response $\mathbf{y}$ satisfying the following autoregression formula:
\begin{align}
    y_L & = x_L \\
    y_i & = \left(\prod_{j>i} [(y_j + x_i) \% (p - 1) + 1] \right) \% p, 1 \leq i < L
\end{align}
This results in every $y_i$ depending on all $y_j, j > i$.
We use $L=20, p=7$ for our experiments, resulting in $N=M=20$.
The size of the training set is $10^6$ while the size of the test set is $10^3$.

\subsubsection{Multiplication}
We generate a multiplication dataset with prompts of the form \verb|x*y=| and responses \verb|z|, where $x \in [10^{19}, 10^{20})$ is a 20-digit positive integer, $y \in [2, 100)$ is a positive integer, and $z=x*y$.
This leads to $N=24,M=22$.
Both $x$ and $y$ are uniformly at random within the respective range.
We generate a training set of size $10^5$ and a test set of size $10^3$.

\subsubsection{Sudoku}
We use the Sudoku dataset from \cite{shah2024causal}.
The prompt is a string of length 81 that contains the flattened initial configuration, where each element is either in $[1, 9]$ denoting a given cell, or $0$ indicating that the value in this cell is missing.
The response is also a string of length 81 that contains the full solution.
$N=M=81$.
The training set contains approximately $1.8*10^6$ samples while the test set contains $10^5$ instances.

\subsubsection{Zebra}
We also use the Zebra dataset from \cite{shah2024causal}.
The prompt for Zebra puzzles consists of a set of clues; we tokenize the special words in the clues with corresponding special tokens.
After tokenization, $N=455,M=42$.
The training set contains about $1.5*10^6$ puzzles while the test set contains $10^5$.

\subsection{Implementation Details}

\subsubsection{Details of ReCOR}

The core algorithm of ReCOR is described in the method section of the main text.
We document additional details below.

The ReCOR-related hyperparameters are listed in Tab.~\ref{tab:hyp_recor}.
For the token query stream and (optional) order query stream, we reuse all parameters of the main stream (including attention projection layers and MLP layers) to ensure that ReCOR always has the same or a smaller parameter count than baselines.
During training, for the sampling of $\boldsymbol{\rho}$, we always greedily sample the position with minimal hardness as predicted by $Q_\theta$, while the token queries $\bar{\boldsymbol{\rho}}$ are sampled using \verb|Exploration mode| in Tab.~\ref{tab:hyp_recor} without duplications.
\verb|Uniform| means uniform sampling $\bar{\boldsymbol{\rho}} \sim \mathcal{U}_\mathcal{A}$, while $\alpha=\ldots$ means using the corresponding entropy coefficient for exploration.
For inference, we always greedily sample both $\rho_t$ and $y_{\rho_t}$.
We use action masking~\cite{huang2020actionmasking} and never pick actions at already filled positions.
We employ $\gamma=0$ across the board, in effect solving a contextual bandit problem.
Since our evaluation datasets require an exact match of all response tokens, the correctness of every token counts, and we do not need the trade-off between short-term and long-term rewards introduced by $\gamma$.
$\gamma=0$ simplifies the learning procedure by eliminating additional value function evaluations while still achieving superior performance, as evidenced in our experiments.
Furthermore, we use the thresholded, sparse reward since this choice slightly improves performance (see additional ablations below), which we also attribute to the exact-match evaluation metric.
Note that the binary cross-entropy loss associated with this reward actually computes an upper bound of the true Q function via Jensen's inequality.
However, since we use $\gamma=0$, the expectation for the Q value concentrates at a single point, rendering the bound tight.

Optionally, we use focal-like techniques~\cite{lin2017focalloss} with focal $\gamma$ listed in Tab.~\ref{tab:hyp_recor} as \verb|Focal coefficient|.
We use focal loss for the RL value loss $\mathcal{L}_\text{SQL}$ instead of the token (language modeling) loss $\mathcal{L}_\text{LM}$, differing from similar techniques used in \cite{ye2025beyond}.
We remark that, as argued in the main text, certain token prediction sub-problems are inherently hard, and we should not expect the model to \textit{solve} them or penalize it for these failures; rather, we simply want the model to \textit{recognize} that these sub-problems are bad through the value loss.

\subsubsection{Hyperparameters}

We list hyperparameters on Autoregression and Multiplication in Tab.~\ref{tab:hyp_arg_mul} while describing ReCOR-related ones on all datasets in Tab.~\ref{tab:hyp_recor}.
Baseline performances for Sudoku and Zebra are reported by \cite{kim2025train}.
In Autoregression and Multiplication, compared with ReCOR, we double the batch size and number of epochs for baselines to match the amount of compute per iteration and number of gradient steps of ReCOR to ensure a fair comparison.

\subsubsection{Other Training Details}

For all experiments, we use mixed-precision training with bfloat16.
The model parameters are kept in full precision (float32).
On Autoregression, Multiplication, and Sudoku, we use the tiny version of GPT-2 with 3 layers, 384 hidden dimensions, and 12 attention heads, resulting in approximately 6M parameters.
On Zebra, we use the nano version with double the depth and parameter count.
Each experiment takes a couple of hours using a single NVIDIA RTX4090 on Autoregression and Multiplication, and less than 2 days using a single NVIDIA A100-80G on Sudoku and Zebra.

\begin{table}
  \caption{Hyperparameters for ReCOR and baselines on Autoregression and Multiplication.}
  \label{tab:hyp_arg_mul}
  \centering
  \begin{tabular}{ccccc}
    \toprule
    Hyperparameter & ReCOR & CLM & AR-GT & (Ada)MDM \\
    \midrule
    \# Model parameters & 6M & 6M & 6M & 6M \\
    Batch size & $1024$ & $2048$ & $2048$ & $2048$ \\
    \# Epochs & $100$ & $200$ & $200$ & $200$ \\
    Optimizer & AdamW & AdamW & AdamW & AdamW \\
    Learning rate & $10^{-3}$ & $10^{-3}$ & $10^{-3}$ & $10^{-3}$ \\
    LR scheduler & Cosine & Cosine & Cosine & Cosine \\
    Weight decay & $0.1$ & $0.1$ & $0.1$ & $0.1$ \\
    Diffusion steps & N/A & N/A & N/A & $20$ \\
    \bottomrule
  \end{tabular}
\end{table}
\begin{table}
  \caption{ReCOR-related hyperparameters on our evaluation datasets.}
  \label{tab:hyp_recor}
  \centering
  \begin{tabular}{ccccc}
    \toprule
    Hyperparameter & ARG & MUL & Sudoku & Zebra \\
    \midrule
    \# Model parameters & 6M & 6M & 6M & 11M \\
    Batch size & $1024$ & $1024$ & $512$ & $512$ \\
    \# Epochs & $100$ & $100$ & $40$ & $50$ \\
    Optimizer & AdamW & AdamW & AdamW & AdamW \\
    Learning rate & $10^{-3}$ & $10^{-3}$ & $3*10^{-4}$ & $3*10^{-4}$ \\
    LR scheduler & Cosine & Cosine & Cosine & Cosine \\
    Weight decay & $0.1$ & $0.1$ & $0.1$ & $0.1$ \\
    Discount factor $\gamma$ & $0.0$ & $0.0$ & $0.0$ & $0.0$ \\
    Focal coefficient & $2.0$ & $2.0$ & $0.0$ & $0.0$ \\
    Threshold $\eta$ & $0.8$ & $0.8$ & $0.8$ & $0.8$ \\
    \# Token queries $K$ & $1$ & $1$ & $8$ & $2$ \\
    \# Order queries $C$ & $0$ & $0$ & $8$ & $2$ \\
    Exploration mode & Uniform & Uniform & $\alpha=0.1$ & $\alpha=0.1$ \\
    \bottomrule
  \end{tabular}
\end{table}

\section{Additional Experiments}

\begin{figure}[t]
  \centering
  \begin{subfigure}{0.48\textwidth}
      \includegraphics[width=\linewidth]{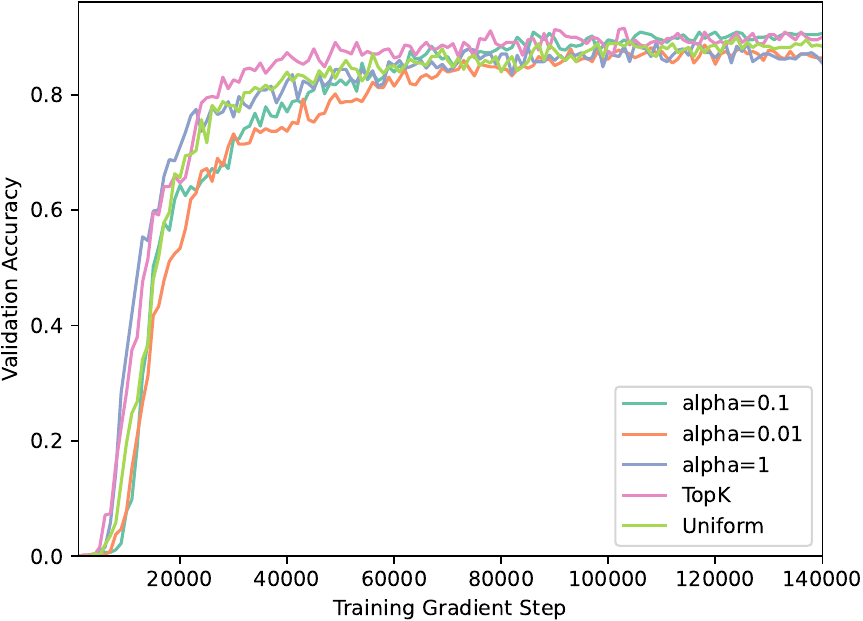}
      \subcaption{Exploration strategy ablation with sparse rewards.}
      \label{fig:abl_exploration}
  \end{subfigure}
  \begin{subfigure}{0.48\textwidth}
      \includegraphics[width=\linewidth]{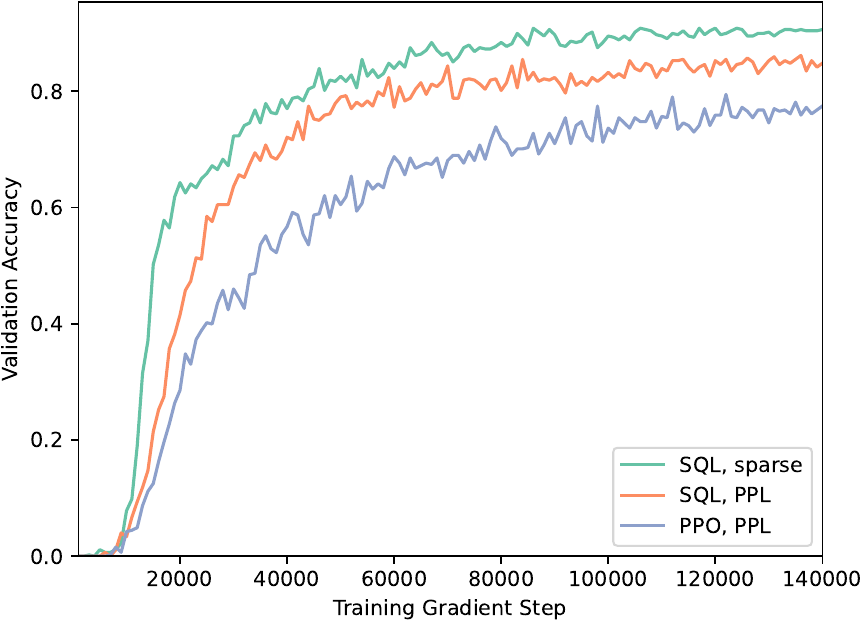}
      \subcaption{Reward function and RL algorithm ablation.}
      \label{fig:abl_rew_rl_alg}
  \end{subfigure}
  
  \caption{Additional ablation experiments regarding the exploration strategy of $\bar{\boldsymbol{\rho}}$ (a) and the choice of reward function and RL algorithm (b).}
  \label{fig:abl_extra}
\end{figure}

\subsection{Exploration Strategies}

We present additional ablation experiments regarding alternative exploration strategies for token queries $\bar{\boldsymbol{\rho}}$ on Sudoku in Fig.~\ref{fig:abl_exploration}.
\verb|alpha=...| corresponds to an entropy-regularized exploration with the designated entropy coefficient, \verb|Uniform| denotes uniform sampling, and \verb|TopK| denotes sampling the $K$ positions with the largest Q-values.
We find ReCOR pretty robust to alternative exploration strategies and opt for $\alpha=0.1$ in our main experiments.

\subsection{Reward Functions and RL algorithms}

Here we study the effects of alternative reward functions and RL algorithms on Sudoku.
We compare our primary setting (Soft Q-learning, sparse reward with threshold $\eta=0.8$) to alternative settings with negated perplexity as rewards and policy-gradient class of RL algorithms.
Note that the sparse reward cannot be used with policy-gradient algorithms.
We use the following loss to implement a PPO~\cite{schulman2017proximal}-like algorithm, replacing $\mathcal{L}_\text{SQL}$:
\begin{equation}
    \mathcal{L}_\text{PPO}(\theta)=\mathbb{E}_{s,a,r,s'} \left[ \frac{\pi_\theta(a \mid s)}{\pi_{\theta_\text{old}}(a \mid s)} \hat{A} - \alpha H(\pi_\theta(\cdot \mid s)) \right]
\end{equation}
where $\hat{A}$ is the batch-normalized advantage.
Note that since ReCOR always stays exactly on-policy and does not reuse actions, $\theta_\text{old}$ is the detached version of the current $\theta$, and the ratio is always $1$, so we do not need the clipping in standard PPO.

As shown in Fig.~\ref{fig:abl_rew_rl_alg}, our value-based choice of RL algorithm with a sparse reward achieves the best overall performance.
Switching to dense rewards still yields decent performance, validating our motivation.
We found a value-based algorithm to slightly outperform policy-gradient methods, potentially due to the fact that it can make full use of return signals while policy-based methods only model the relative difference between actions.

\end{document}